%% file: main.tex
\documentclass[journal]{IEEEtran}

\usepackage{amsmath,amsfonts}
\usepackage{array}
\usepackage[caption=false,font=normalsize,labelfont=sf,textfont=sf]{subfig}
\usepackage{textcomp}
\usepackage{stfloats}
\usepackage{verbatim}

\usepackage{microtype}
\usepackage{booktabs} 
\usepackage{algpseudocode}
\usepackage{array}
\usepackage{url}

\usepackage{mymacros}

\usepackage{hyperref}       
\usepackage{nicefrac}       
\usepackage{xcolor}         
\usepackage{amsthm}         
\usepackage{multicol}
\usepackage[pdftex]{graphicx}
\usepackage[linesnumbered,ruled,vlined]{algorithm2e}  
\usepackage{bbm}
\usepackage{caption} 
\newtheorem{theorem}{Theorem} 

\newcommand{\gda}{\textit{ GD\textsubscript{a}}}
\newcommand{\gdr}{\textit{GD\textsubscript{r}}}
\newcommand{\ltc}{\textit{L2C}}
\newcommand{\projectwebsite}{\url{https://mit-realm.github.io/radium}}

\begin{document}
%
\title{RADIUM: Predicting and Repairing End-to-End Robot Failures using Gradient-Accelerated Sampling}
%
%
%

\author{Charles~Dawson,~\IEEEmembership{Member,~IEEE,}
        Anjali~Parashar,~\IEEEmembership{Member,~IEEE,}
        and~Chuchu~Fan,~\IEEEmembership{Member,~IEEE}
\thanks{C. Dawson and C. Fan are with the Department
of Aeronautics and Astronautics, A. Parashar is with the Department of Mechanical Engineering, MIT, Cambridge,
MA, 02139 USA e-mail: \texttt{\{cbd, anjalip, chuhchu\}@mit.edu}.}
}

\markboth{Transactions on Robotics}%
{Dawson \MakeLowercase{\textit{et al.}}: Predicting and Repairing End-to-End Robot Failures}

\maketitle


\begin{abstract}
Before autonomous systems can be deployed in safety-critical applications, we must be able to understand and verify the safety of these systems. For cases where the risk or cost of real-world testing is prohibitive, we propose a simulation-based framework for a) predicting ways in which an autonomous system is likely to fail and b) automatically adjusting the system's design and control policy to preemptively mitigate those failures. Existing tools for failure prediction struggle to search over high-dimensional environmental parameters, cannot efficiently handle end-to-end testing for systems with vision in the loop, and provide little guidance on how to mitigate failures once they are discovered.
We approach this problem through the lens of approximate Bayesian inference and use differentiable simulation and rendering for efficient failure case prediction and repair. For cases where a differentiable simulator is not available, we provide a gradient-free version of our algorithm, and we include a theoretical and empirical evaluation of the trade-offs between gradient-based and gradient-free methods. We apply our approach on a range of robotics and control problems, including optimizing search patterns for robot swarms, UAV formation control, and robust network control. Compared to optimization-based falsification methods, our method predicts a more diverse, representative set of failure modes, and we find that our use of differentiable simulation yields solutions that have up to 10x lower cost and requires up to 2x fewer iterations to converge relative to gradient-free techniques. In hardware experiments, we find that repairing control policies using our method leads to a 5x robustness improvement. Accompanying code and video can be found at \projectwebsite{}.
\end{abstract}


%
\IEEEpeerreviewmaketitle

\section{Introduction}

\IEEEPARstart{F}{rom} aerial robots to transportation and logistics systems and power grids, autonomous systems play a central, and often safety-critical, role in modern life. Even as these systems grow more complex and ubiquitous, we have already observed failures in autonomous systems like autonomous vehicles and power networks resulting in the loss of human life. Given this context, it is important that we be able to verify the safety of autonomous systems \textit{prior} to deployment; for instance, by understanding the different ways in which a system might fail and proposing repair strategies.

Human designers often use their knowledge of likely failure modes to guide the design process; indeed, systematically assessing the risks of different failures and developing repair strategies is an important part of the systems engineering process. However, as autonomous systems grow more complex, it becomes increasingly difficult for human engineers to manually predict likely failures.

Adversarial testing methods propose to solve this problem by searching for counterexamples where the learned system performs poorly, then retraining on those counterexamples~\cite{madryDeepLearningModels2018, salmanUnadversarialExamplesDesigning2021,hanselmannKINGGeneratingSafetyCritical2022a,dingLearningCollideAdaptive2020a,corsoAdaptiveStressTesting2019,wangAdvSimGeneratingSafetyCritical2021}. Adversarial methods are typically greedy, using gradient-based or gradient-free optimization to seek out the most severe or most likely failures, but this leads to a critical issue: a loss of diversity in the counterexamples. If the counterexamples over-represent certain cases, then the retraining process will over-fit to those cases, reducing robustness.

The challenge of finding diverse counterexamples has motivated recent work in \textit{rare-event prediction} using methods like Markov Chain Monte Carlo (MCMC) and importance sampling~\cite{sinhaNeuralBridgeSampling2020,deleckiModelbasedValidationProbabilistic2023a,zhouRoCUSRobotController2021,okellyScalableEndtoEndAutonomous2018,corsoAdaptiveStressTesting2019}. Unfortunately, existing rare-event prediction methods, particularly importance sampling, suffer from the curse of dimensionality as rare events become sparse in high-dimensional search spaces~\cite{betancourtConceptualIntroductionHamiltonian2017}. Moreover, existing failure prediction methods provide little guidance on how to update the policy once failures have been discovered.

\begin{figure*}[t]
  \centering
  \includegraphics[width=\linewidth]{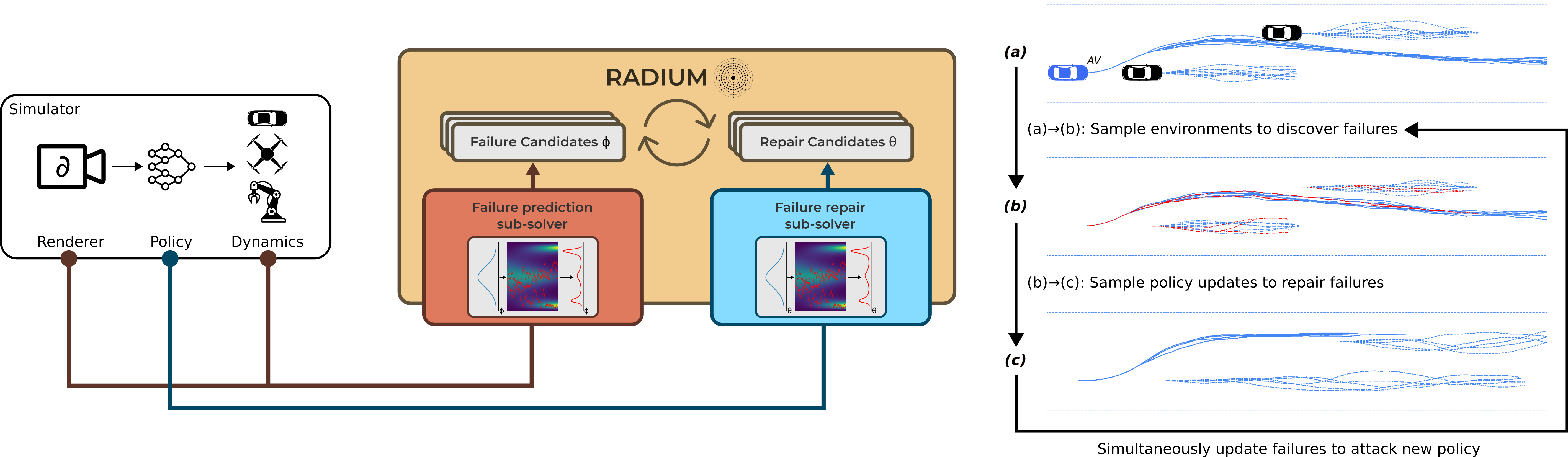}
  \caption{An overview of our approach for closed-loop rare-event prediction, which efficiently predicts and repairs failures in autonomous systems. Our framework alternates between failure prediction and repair sub-solvers, which use a simulated environment to efficiently sample from the distributions~\eqref{eq:failure_logprob_basic} and~\eqref{eq:repair_logprob_basic}. We use differentiable rendering and simulation to accelerate our method with end-to-end gradients, but we also propose a gradient-free implementation.}\label{fig:architecture}
\end{figure*}

In this paper, we aim to close the gap between adversarial training and rare-event prediction with RADIUM: a framework that simultaneously predicts diverse, challenging failures and updates the control policy to repair those failures, as shown in Fig.~\ref{fig:architecture}. To efficiently explore the failure space, we start with highly likely failures and gradually expand our search to more severe rare counterexamples, continuously repairing the policy as the failure distribution shifts. We make the following contributions:

\begin{enumerate}
    \item We reframe adversarial optimization as a sequential inference problem, leading to a novel framework for predicting and repairing a diverse set of failures.
    \item We develop both gradient-free and gradient-based variants of our framework, relying on differentiable simulation and rendering for the latter, and evaluate the performance trade-offs when scaling to high-dimensional problems.
    \item We provide a theoretical analysis of our sequential inference framework, proving correctness and asymptotic convergence for both variants. For the gradient-based variant, we also provide finite-sample convergence rates in a restricted setting.
\end{enumerate}

We demonstrate our approach through extensive benchmarking in simulation, demonstrating state-of-the-art performance on a range of robotics and control problems. We also include hardware experiments showing sim2real transfer of both predicted failure modes and repaired policies, showing an up to 5x improvement in the safety of a vision-in-the-loop control policy.

\subsection{Differences from conference version}

This paper extends our results in~\cite{dawsonBayesianApproachBreaking2023} with three additional contributions:
\begin{enumerate}
    \item \textbf{Vision-in-the-loop:} We extend our approach to include efficient gradient-based testing for systems with vision-in-the-loop. Visual perception is an important component of modern robots, but prior works have been unable to efficiently test and repair these systems due to the lack of gradients through the vision pipeline. We address this gap by developing a differentiable rendering-based pipeline for failure prediction and repair.
    \item \textbf{Theoretical results:} We provide an additional theorem characterizing the stationary distribution of our sequential inference framework, providing insight into the connections between our inference-based approach and traditional adversarial optimization.
    \item \textbf{Additional simulation studies:} We include additional simulation studies for vision-in-the-loop testing, including an additional baseline method.
    \item \textbf{Additional hardware demonstrations:} We provide additional hardware demonstrations of vision-in-the-loop testing of an autonomous 1/10th-scale car.
\end{enumerate}

\section{Prior work}

Our work builds on a rich literature of different verification and testing techniques, which we review here.

\paragraph{Model-based verification}

Early work on model based verification relied on logical models to search for failures using satisfiability (SAT) solvers~\cite{dekleerDiagnosingMultipleFaults1987,benardRemoteAgentExperiment2000}. The computational expense of SAT, which prevented these methods from scaling to high-dimensional problems, motivates the development of more recent approaches using mathematical dynamics models and tools like reachability analysis~\cite{annpureddySTaLiRoToolTemporal2011a} and optimal control~\cite{chouUsingControlSynthesis2018} to identify counterexamples. The challenge in applying all of these methods is that it is often difficult (or impossible) to construct a symbolic model of the system under test (e.g. when vision is involved). In this work, our aim is to preserve the interpretability of model-based verification without relying on a symbolic model. Instead, we develop a simulation-based approach using automatic differentiation (when available) to accelerate the search for counterexamples. Simulators are widely used in robotics, and recent work on differentiable simulation and rendering has made gradients available for a range of scenarios~\cite{dawsonRobustCounterexampleguidedOptimization2022b, amosOptNetDifferentiableOptimization2017,belubute_peres_lcp_physics,jakobDRJITJustintime2022,jainAnalyzingImprovingNeural2020,huDiffTaichiDifferentiableProgramming2019,zhongGuidedConditionalDiffusion2022,dawsonCertifiableRobotDesign2022}.

\paragraph{Adversarial testing}

A common approach in the falsification literature is adversarial optimization, which formulates the search for counterexamples as a two-player game between the system designer and the environment~\cite{dawsonRobustCounterexampleguidedOptimization2022b,dontiAdversariallyRobustLearning2021,yaghoubiGrayboxAdversarialTesting2018,corsoSurveyAlgorithmsBlackBox2021,xuSafeBenchBenchmarkingPlatform2022,riedmaierSurveyScenarioBasedSafety2020,okellyScalableEndtoEndAutonomous2018,corsoAdaptiveStressTesting2019,wangAdvSimGeneratingSafetyCritical2021,sunCornerCaseGeneration2021,zhongGuidedConditionalDiffusion2022,corsoInterpretableSafetyValidation2020a,zhangAdversarialRobustnessTrajectory2022,hanselmannKINGGeneratingSafetyCritical2022a}. Adversarial methods have been developed in both model-based and model-free contexts, where the main distinction is that the former assume access to gradients. Gradient-based adversarial methods have demonstrated impressive sample efficiency~\cite{dawsonRobustCounterexampleguidedOptimization2022b,dontiAdversariallyRobustLearning2021}, but the drawback is that they rely on local gradient ascent to search for counterexamples, typically yielding only a single adversarial counterexample that may be only locally optimal. Moreover, existing gradient-based methods cannot handle systems with visual feedback. \cite{hanselmannKINGGeneratingSafetyCritical2022a} find that skipping the rendering step during backpropagation is sufficient for gradient-based failure-case generation, but this method is not able to repair visual-feedback policies. \cite{sinhaNeuralBridgeSampling2020} learn a proxy model for the end-to-end dynamics, including a vision-based policy, but end-to-end proxy modeling is generally not applicable to rare-event generation, since it is difficult to include enough failure examples in the training data for the proxy model.
On the other hand, gradient-free methods reduce the risk of getting stuck in local optima and support vision in the loop, but they require additional computation and struggle to scale to high-dimensional search spaces. In contrast, in this work we develop a probabilistic gradient-based approach, using sample-efficient gradient-based sampling algorithms to avoid local minima and scale to high dimensional search spaces, and using differentiable rendering to support visual feedback systems. We also provide a gradient-free variant of our method for use in cases when a differentiable simulator is not available.

\paragraph{Inference}

Our probabilistic approach builds off of prior work on inference as a verification tool. \cite{okellyScalableEndtoEndAutonomous2018} propose an end-to-end verification tool for autonomous vehicles based on gradient-free adaptive importance sampling. \cite{zhouRoCUSRobotController2021} use gradient-free Markov Chain Monte Carlo (MCMC) to find counterexamples. \cite{sinhaNeuralBridgeSampling2020} and \cite{deleckiModelbasedValidationProbabilistic2023a} use gradient-based MCMC to estimate the risk of failure and find counterexamples, respectively.
These prior works are focused only on predicting failure modes; they are not able to improve the system (e.g. by re-optimizing the controller) to fix these failures once they have been discovered. Our method aims to fill this gap by combining failure mode prediction and repair, exploiting the duality between these problems to efficiently search for both a diverse set of counterexamples and updates to the system's policy or design that reduces the severity of those failures.

\section{Problem statement}

We begin by considering a general autonomous system that receives observations $o \in \mathcal{O}$ and makes control decisions using a learned control policy $\pi_\theta: \mathcal O \mapsto \mathcal A$, where $\theta$ are the parameters of that policy. This agent operates in closed loop with an environment with dynamics $f_{\phi}: \mathcal X \mapsto \mathcal X$ and rendering function $R_{\phi}: \mathcal X \mapsto \mathcal O$, where uncertainty in the environment manifests as uncertain parameters $\phi$ with probability density $p_{\phi, 0}$. Without loss of generality, we assume that randomness in the environment has been ``factored out'' into $\phi$ (so that $f$ and $R$ are deterministic given $\phi$). We assign a cost $J(\theta, \phi)$ to a pair of policy and environmental parameters by rolling out for a fixed $T$-step horizon:
\begin{align}
    o_t &= R_{\phi}(x_t) \label{eq:rendering}\\
    x_{t+1} &= f_{\phi}(x_t, \pi_\theta(o_t)) \label{eq:dynamics}\\
    J(\theta, \phi) &= J(x_0, \ldots, x_T) \label{eq:cost}
\end{align}

In this context, \textit{failure prediction} involves finding environmental parameters $\phi$ that induce high cost for given policy parameters $\theta$, i.e. finding multiple solutions $\phi^*(\theta) = \text{find}_\phi J(\theta, \phi) \geq J^*$ for failure threshold $J^*$,
while \textit{failure repair} involves modifying the initial policy parameters $\theta_0$ to achieve low costs despite possible variation in $\phi$, i.e. finding a nearby $\theta^* = \min_\theta ||\theta_0 - \theta||^2 \text{ s.t. } E_\phi[J(\theta,\phi)] \leq J^*$. With slight abuse of notation, we will use $J(\theta, \phi)$ to refer to the composition of the simulator, renderer, and cost function.

\subsection{Failure prediction}

Simply optimizing for the highest-severity or most-likely failure will give an incomplete picture of the system's performance and lead to more conservative behavior. Instead, we balance the prior likelihood of a disturbance with the severity of the induced failure by sampling failures from the pseudo-posterior
\begin{equation}
    p_{\rm{failure}}(\phi; \theta) \propto p_{\phi, 0}(\phi)e^{-[J^* - J(\theta, \phi)]_+} \label{eq:failure_logprob_basic}
\end{equation}
where $J^*$ is the cost threshold for a failure event and $[\cdot]_+$ is the exponential linear unit. Intuitively, we can interpret this likelihood as a posterior over environmental parameters $\phi$ conditioned on a failure occurring \cite{sinhaNeuralBridgeSampling2020,zhouRoCUSRobotController2021,maSamplingCanBe2019,dawsonBayesianApproachBreaking2023}. By framing the search for failures prediction as a sampling problem, rather than the traditional adversarial optimization, we gain a number of advantages. First, we are able to generate a more diverse set of failure modes than would be discovered by an extremum-seeking approach, as shown by our experiments in Section~\ref{experiments}. Second, we are able to draw on a rich literature of theoretically well-motivated sampling algorithms to develop our approach, as we discuss in Sections~\ref{approach} and~\ref{theory}.

\subsection{Failure repair}
This is not the first paper to take a sampling-based approach to failure prediction; for example, \cite{okellyScalableEndtoEndAutonomous2018}, \cite{sinhaNeuralBridgeSampling2020}, and \cite{zhouRoCUSRobotController2021} also approach failure prediction using this lens. Our insight is that this sampling framework can be extended to not only predict failures but also repair the underlying policy, thus mitigating the impact of the failure.
Given initial policy parameters $\theta_0$ and a population of anticipated failure modes $\phi_1, \ldots, \phi_n$, we can increase the robustness of our policy by sampling from a corresponding repair pseudo-posterior, similar to Eq.~\eqref{eq:failure_logprob_basic},
\begin{equation}
p_{\rm{repair}}(\theta; \phi_1, \ldots, \phi_n) \propto p_{\theta, 0}(\theta; \theta_0)e^{-\sum_{\phi_i}[J(\theta, \phi_i) - J^*]_+ / n} \label{eq:repair_logprob_basic}
\end{equation}
where the prior likelihood $p_{\theta, 0}$ regularizes the search for repaired policies that are close to the original policy.
Intuitively, this distribution of repaired policies can be seen as a posterior over policies conditioned on the event that a failure \textit{does not} occur in the given scenarios. Sampling from this posterior can be seen as a form of regularized re-training on the set of predicted failures, since maximizing the log of~\eqref{eq:repair_logprob_basic} is equivalent to minimizing the empirical risk $\sum_{\phi_i}[J(\theta, \phi_i) - J^*]_+ / n$ with regularization $||\theta-\theta_0||_2^2$ (assuming a Gaussian prior). This connection helps motivate our use of~\eqref{eq:repair_logprob_basic}, but we find empirically in Section~\ref{experiments} that the increased diversity from sampling rather than policy gradient optimization yields better policies in practice.

\section{Approach}\label{approach}

Previous works have shown that sampling from a failure distribution like Eq.~\eqref{eq:failure_logprob_basic} can generate novel failures~\cite{zhouRoCUSRobotController2021,sinhaNeuralBridgeSampling2020,deleckiModelbasedValidationProbabilistic2023a}, but several challenges have prevented these works from considering end-to-end policy repair as well. Our main contribution is a framework for resolving these challenges and enabling simultaneous failure prediction and repair, which we call \textit{RADIUM} (Robustness via Adversarial Diversity using MCMC, illustrated in Fig.~\ref{fig:architecture}). We have designed this framework to take advantage of problem structure (e.g. differentiability) when possible, but we provide the ability to swap gradient-based subsolvers for gradient-free ones when needed, and we include a discussion of the associated trade-offs.

\paragraph{Challenge 1: Distribution shift during retraining} Previous methods have proposed generating failure examples for use in retraining, but there is an inherent risk of distribution shift when doing so. Once we repair the policy, previously-predicted failures become stale and are no longer useful for verification (i.e. the distribution of likely failures has shifted). In the worst case, this can lead to overconfidence if the system claims to have repaired all previously-discovered failures while remaining vulnerable to other failures. To address this issue, we interleave failure and repair steps to continuously update the set of predicted failures as we repair the policy, creating an adversarial sampling process that generates a robust repaired policy along with a set of salient failure cases.

\paragraph{Challenge 2: Exploring diverse failure modes} Traditional methods like Markov chain Monte Carlo (MCMC) are able to sample from non-normalized likelihoods like~\eqref{eq:failure_logprob_basic} and \eqref{eq:repair_logprob_basic}, but they struggle to fully explore the search space when the likelihood is highly multi-modal. To mitigate this issue, we take inspiration from the recent success of diffusion processes \cite{songScoreBasedGenerativeModeling2023,zhongGuidedConditionalDiffusion2022} and sequential Monte Carlo algorithms \cite{rubinoIntroductionRareEvent2009a} that interpolate between an easy-to-sample prior distribution and a multi-modal target distribution. Instead of sampling directly from the posterior, we begin by sampling from the unimodal, easy-to-sample prior and then smoothly interpolate to the posterior distributions~\eqref{eq:failure_logprob_basic}-\eqref{eq:repair_logprob_basic}. This process yields the tempered likelihood functions:
\begin{align}
    \tilde{p}_{\rm{failure}} &\propto p_{\phi, 0}(\phi)e^{-\tau[J^* - J(\theta, \phi)]_+} \label{eq:failure_logprob_tempered} \\
    \tilde{p}_{\rm{repair}} &\propto p_{\theta, 0}(\theta, \theta_0)e^{-\frac{\tau}{n}\sum_{\phi_i}[J(\theta, \phi_i)-J^*]_+} \label{eq:repair_logprob_tempered}
\end{align}
where the tempering parameter $\tau$ is smoothly varied from $0$ to $1$. When $\tau = 0$, this is equivalent to sampling from the prior distributions, and when $\tau \to 1$ we recover the full posteriors~\eqref{eq:failure_logprob_basic}-\eqref{eq:repair_logprob_basic}. This tempering process reduces the risk of overfitting to one particular mode of the failure distribution and encourages even exploration of the failure space.

\paragraph{Challenge 3: Efficiently sampling in high dimension}
Previous works have proposed a wide variety of sampling algorithms that might be used as sub-solvers in our framework, including MCMC methods like random-walk Metropolis-Hastings (RMH;~\cite{hastingsMonteCarloSampling1970}), Hamiltonian Monte Carlo (HMC;~\cite{nealMCMCUsingHamiltonian2011}), and the Metropolis-adjusted Langevin algorithm (MALA;~\cite{julianbresagCommentsGrenadierMiller1994}), variational inference methods like Stein Variational Gradient Descent (SVGD;~\cite{liuSteinVariationalGradient2016a}), and other black-box methods like adaptive importance sampling~\cite{okellyScalableEndtoEndAutonomous2018}. RADIUM is able to use any of these sampling methods as sub-solvers for either the prediction or repair. Generally, these sampling methods can be classified as either gradient-free or gradient-based. Theoretical and empirical evidence suggests that gradient-based methods can enjoy faster mixing time in high dimensions on certain classes of sufficiently smooth (but non-convex) problems~\cite{maSamplingCanBe2019}, but autonomous systems with visual feedback have historically  been treated as black-boxes due to an inability to backpropagate through the rendering step \cite{zhouRoCUSRobotController2021,okellyScalableEndtoEndAutonomous2018,sinhaNeuralBridgeSampling2020}.
To enable the use of gradient-based samplers in RADIUM, we draw upon recent advances in differentiable simulation and rendering~\cite{huDiffTaichiDifferentiableProgramming2019,lelidecDifferentiableRenderingPerturbed2021} provide end-to-end gradients. In Sections~\ref{theory} and~\ref{experiments}, we provide theoretical and empirical evidence of a performance advantage for gradient-based samplers, but in order to make RADIUM compatible with existing non-differentiable simulators we also conduct experiments where RADIUM uses gradient-free sampling subroutines. We provide more discussion of differentiable simulation and rendering methods in Section~\ref{diffsim}.

\subsection{RADIUM}

Pseudocode for RADIUM is provided in Algorithm~\ref{alg:adv_diffusion}. The algorithm maintains separate populations of candidate repaired policies $[\theta_1, \ldots, \theta_n]$ and failures $[\phi_1, \ldots, \phi_n]$ that are updated over $N$ sampling rounds. In each round, we sample a set of new candidate policies from the repair generating process~\eqref{eq:repair_logprob_tempered}, then sample a new set of failures that attack the current population of policies. In practice, we average the tempered failure log probability~\eqref{eq:failure_logprob_tempered} over the population of candidate designs, which results in a smoother distribution.

RADIUM supports a wide range of subroutines for sampling candidate failures and repaired policies. In our experiments and the provided implementation, we include RMH and MALA (gradient-free and gradient-based MCMC algorithms, respectively); we choose these particular methods to provide a direct comparison between similar algorithms with and without gradients (MALA reduces to RMH when the gradient is zero). Pseudocode for MALA is included in Algorithm~\ref{alg:mala}.


\begin{algorithm}
\SetAlgoLined
\SetKwInput{Input}{Input}
\SetKwInput{Return}{Return}
\Input{$N$ rounds, $K$ steps per round, stepsize $\lambda$, population size $n$, tempering rate $\alpha$, sampling algorithm (e.g. MALA as in Alg.~\ref{alg:mala})}

Sample initial failures and policies using priors: $[\phi_1, \ldots, \phi_n]_0 \overset{\mathrm{iid}}{\sim} p_{\phi, 0},\ [\theta_1, \ldots, \theta_n]_0 \overset{\mathrm{iid}}{\sim} p_{\theta, 0}$\;

\For{$i = 1, \ldots, N$}{
    $\tau \gets 1 - e^{-\alpha i / N}$\tcp*{Tempering schedule}
    Sample $[\theta_1, \ldots, \theta_n]_i \overset{\mathrm{iid}}{\sim}$ \eqref{eq:repair_logprob_tempered}\tcp*{Sample repaired policies}\label{line:policy_update1}
    %
    %
    Sample $[\phi_1, \ldots, \phi_n]_i \overset{\mathrm{iid}}{\sim}$ \eqref{eq:failure_logprob_tempered}\tcp*{Generate failures attacking $\theta^*_i$}\label{line:env_update}
}

\Return{Repaired policy $\theta^*_N = \argmax_i$ \eqref{eq:repair_logprob_tempered} and failures $[\phi_1, \ldots, \phi_n]_N$ attacking that policy.}

\caption{RADIUM: Robustness via Adversarial Diversity using MCMC}\label{alg:adv_diffusion}
\end{algorithm}

\begin{algorithm}
\caption{Metropolis-adjusted Langevin algorithm\label{alg:mala}}
\DontPrintSemicolon
    \SetAlgoLined
    \SetKwInput{Input}{Input}
    \SetKwInput{Return}{Return}
    \Input{Initial $x_0$, steps $K$, stepsize $\epsilon$, density $p(x)$.}
    \For{$i = 1, \ldots, K$}
    {
        Sample $\eta \sim \cN(0, 2\epsilon I)$ \Comment{Gaussian noise}\;
        $x_{i+1} \gets x_i + \epsilon \nabla \log p(x_i) + \eta$ \Comment{Propose next state}\;\label{mala:step}
        $P_{accept} \gets \frac{p(x_{i+1}) e^{-||x_i - x_{i+1} - \epsilon \nabla \log p(x_{i+1})||^2 / (4\epsilon)}}{p(x_{i}) e^{-||x_{i+1} - x_{i} - \epsilon \nabla \log p(x_{i})||^2 / (4\epsilon)}}$ \;
        With probability $1 - \min(1, P_{accept})$:\;
        \hspace{2em}$x_{i+1} \gets x_{i}$ \Comment{Accept/reject proposal}\;\label{mala:mh}
    }
    \Return{$x_K$ (approximately sampled from $p(x)$)}
\end{algorithm}

\subsection{Differentiable simulation and rendering}\label{diffsim}

A key component of our approach is making use of gradients from differentiable simulation and rendering, when available. Although we provide a gradient-free version of RADIUM for cases when a differentiable environment is unavailable, our empirical results in Section~\ref{experiments} find that gradients can substantially improve sample efficiency in many domains. This section provides additional details on differentiable environments with both simulation and rendering.

Recall that our system consists of dynamics $f_\phi$, a rendering function $R_\phi$, and a cost $J$. In this section, we consider the case where all three functions (and the policy $\pi_\theta$) are differentiable almost everywhere and discuss how the gradient $\nabla J$ may be computed. Applying the chain rule, we see that this will require backpropagating through the dynamics, rendering function, and policy:
\begin{align}
    \nabla_{(\theta, \phi)} J &= \sum_{t=0}^T \nabla_{x_t} J \cdot D_{(\theta, \phi)} x_t \\
    D_{(\theta, \phi)} x_0 &= \mathbf{0} \\
    D_{\theta} x_{t+1} =& D_{x_t}f_\phi(x_t, u_t) \cdot D_\theta x_t + D_{u_t}f_\phi(x_t, u_t) \cdot \left( D_\theta \pi_\theta (o_t) \right. \nonumber\\
    & \left. + D_{o_t} \pi_\theta (o_t) \cdot D_{x_t} R(x_t) \cdot D_\theta x_t \right) \\
    D_{\phi} x_{t+1} =& D_\phi f_\phi(x_t, u_t) + D_{x_t}f_\phi(x_t, u_t) \cdot D_\phi x_t \\
    &+ D_{u_t}f_\phi(x_t, u_t) \cdot D_{o_t} \pi_\theta (o_t) D_{x_t} R(x_t) \cdot D_\phi x_t \nonumber
\end{align}
where we use the shorthand $u_t = \pi_\theta(o_t)$ and $D_x f(x, y)$ to denote the Jacobian of a function $f$ with respect to $x$ evaluated at $(x, y)$. To avoid computing this Jacobian by hand, we instead implement $f$, $R$, and $J$ using a framework like JAX or PyTorch and compute the gradient $\nabla_{(\theta, \phi)} J$ using reverse-mode automatic differentiation. In most cases, $J$ is a relatively simple function of $x_t$, and so $\nabla_{x_t} J$ can be evaluated easily, but additional care is required for the dynamics $f$ and rendering function $R$.

When backpropagating through the dynamics, there are two possible issues that must be handled: numerical integration and implicit dynamics. Numerical integration is used to convert continuous-time dynamics to discrete time. Different simulators make use of different numerical integration algorithms, ranging from simple first-order Euler methods to higher-order methods like the Runge-Kutte algorithm. In many cases, it is sufficient to simply implement the integration algorithm in an automatically differentiable framework like JAX, which yields accurate gradients at the expense of high memory usage during backpropagation; when memory usage is a concern, alternatives like the adjoint method may be used~\cite{kidgerNeuralDifferentialEquations2022}.

The second possible issue that may arise when differentiating $f$ are implicit dynamics. Many applications, including robotic manipulation, do not represent the dynamics in closed form $x_{t+1} = f(x_t, u_t)$, instead relying on an implicit form $f(x_{t+1}, x_t, u_t) = 0$. In these cases, evaluating the dynamics requires solving an optimization problem at each step. Although it is technically possible to automatically differentiate the numerical solution of this optimization, doing so will usually result in poor-quality gradients; instead, we can use the implicit function theorem to back-propagate through implicit dynamics~\cite{amosOptNetDifferentiableOptimization2017}. The development of off-the-shelf differentiable simulators is an active area of research, with differentiable implementations of contact dynamics~\cite{belubute_peres_lcp_physics}, soft materials~\cite{hu2019chainqueen}, and fluid dynamics~\cite{lee2023aquarium}, to name just a few.

\begin{figure}
    \centering
    \includegraphics[width=\linewidth]{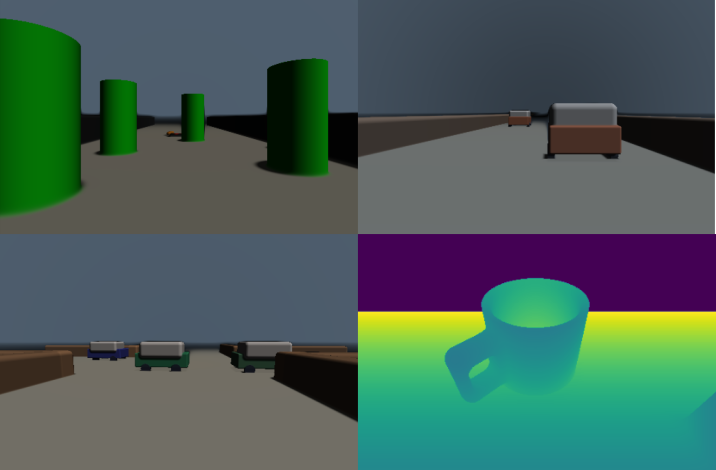}
    \caption{Example images rendered using our basic differentiable rendering engine. Bottom-right shows a depth-only image; the rest are RGB images with directional lighting.}
    \label{fig:example_rendering}
\end{figure}

When the system under test includes visual feedback, we must additionally backpropagate through the rendering step, where the primary difficulty comes from differentiating through discontinuities caused by occlusion. The interested reader is referred to the tutorial in~\cite{zhaoPhysicsbasedDifferentiableRendering2020}. Recent years have seen increasing development of differentiable rendering algorithms and accompanying software frameworks~\cite{jakobDRJITJustintime2022,Mitsuba3}. While there is not yet a readily-available unified environment that combines differentiable simulation and rendering, we hope that this gap will be filled in coming years. In this work, we implement a custom simulation and rendering environment for our benchmark problems; this environment is available at \projectwebsite{}, but we look forward to the development and deployment of increasingly realistic differentiable environments. Our implementation uses signed distance functions to represent geometry, renders the image using raytracing, and approximates the gradients of the raytracing step using implicit differentiation. Example images rendered using our engine is shown in Fig.~\ref{fig:example_rendering}.

\section{Theoretical analysis}\label{theory}

The iterative adversarial sampling process defined in Alg.~\ref{alg:adv_diffusion} raises a few theoretical questions. First, when can we expect the individual sampling steps on lines~\ref{line:policy_update1} and~\ref{line:env_update} to converge, and under what conditions might we expect a gradient-based sampling sub-routine to converge faster than a gradient-free one? Second, assuming that these individual samplers converge, what sort of policies will result from the adversarial sampling loop in Alg.~\ref{alg:adv_diffusion}?

\paragraph{Convergence and gradient acceleration} RADIUM inherits the asymptotic convergence guarantees of the particular subsolvers used for each sampling step. For example, when using an MCMC sampler, so long as that sampler can propose arbitrarily large steps with non-zero probability and satisfies detailed balance (e.g. through the use of a Metropolis adjustment), then the sampler will produce samples asymptotically close to the target sampling distribution. Since the conditions for asymptotic convergence of MCMC samplers are relatively weak~\cite{hastingsMonteCarloSampling1970}, it is more interesting to ask about finite-sample convergence rates; in particular, under what conditions can we expect gradient-based samplers like MALA to accelerate convergence to the target distribution?

In many robotics problems, even when analytical gradients are available, it is unclear whether these gradients are useful for optimization (i.e. low empirical bias and variance;~\cite{suhDifferentiableSimulatorsGive2022}). Here, we build on recent theoretical results by~\cite{maSamplingCanBe2019} to provide sufficient conditions for fast, polynomial-time convergence of gradient-based samplers in our setting.

\begin{theorem}\label{thm:convergence}
    Let $J \circ S$ be a $L$-Lipschitz smooth cost function (i.e. $\nabla J \circ S$ is $L$-Lipschitz continuous), let the log prior distributions $\log p_{\phi ,0}$ and $\log p_{\theta,0}$ be Lipschitz smooth everywhere and $m$-strongly convex outside a ball of finite radius $R$, and let $d = \max\pn{\dim{\theta}, \dim{\phi}}$ be the dimension of the search space. If $m > L$, then MALA with appropriate step size will yield samples within $\epsilon$ total variation distance of the target distributions~\eqref{eq:failure_logprob_tempered} and~\eqref{eq:repair_logprob_tempered} with total number of sampling steps $\leq \widetilde{\mathcal{O}}\pn{d^2 \ln\frac{1}{\epsilon}}$.
\end{theorem}

A proof is given in the appendix, which also provides the required step size for the MALA sampler. The key idea of the proof is to rely on the log-concavity of the prior distributions to dominate the non-convexity of the cost function sufficiently far from the central modes. Theorem~\ref{thm:convergence} requires smoothness assumptions on the cost; we recognize that this assumption is difficult to verify in practice and does not hold in certain domains (notably when rigid body contact is involved). However, in the problems we consider it is possible to smooth both the renderer and scene representation (by blurring the scene and using smooth signed distance functions), thus smoothing the gradients of $J$. The smoothness and convexity conditions hold for many common prior distributions, such as Gaussian and smoothed uniform distributions.

\paragraph{Adversarial Joint Distribution} Even if the samplers for both policy and environmental parameters converge within each round of Alg.~\ref{alg:adv_diffusion}, it is not clear what will be the effect of running these samplers repeatedly in an adversarial manner. Our next theoretical result defines the joint distribution of $\theta$ and $\phi$ as a result of this adversarial sampling loop. To simplify the theoretical analysis, we consider the case when population size $n=1$, and we replace the smooth ELU with a ReLU in~\eqref{eq:failure_logprob_basic} and~\eqref{eq:repair_logprob_basic}.

\begin{theorem}\label{thm:joint}
    The iterative adversarial sampling procedure in Alg.~\ref{alg:adv_diffusion} yields policies drawn from a marginal distribution with density function
    \begin{align}
        f_\theta(\theta^*) = p_{\theta, 0}(\theta^*) &\left( \frac{\expectation_{\phi \sim p_{\phi, 0}}\left[ e^{J(\theta^*, \phi) - J^*} | J(\theta^*, \phi) \leq J^* \right]}{\expectation_{\phi \sim p_{\phi, 0}}\left[ e^{J(\theta^*, \phi) - J^*} \right]} \right. \nonumber\\
        &\qquad + \left.\frac{\mathbb{P}[J(\theta^*, \phi) > J^*]}{\expectation_{\phi \sim p_{\phi, 0}}\left[ e^{J(\theta^*, \phi) - J^*}\right]}\right)\label{eq:marginal}
    \end{align}
    where $\mathbb{P}(J(\theta^*, \phi) > J^* = \expectation_{\phi \sim p_{\phi, 0}}[\mathbbm{1}(J(\theta^*, \phi) \geq J^*)]$ is the probability of failure when $\phi$ is sampled from the prior distribution.
\end{theorem}

The proof is included in the appendix and follows from the Hammersley-Clifford theorem for Gibbs samplers~\cite{robertMonteCarloStatistical2004}. The first term in the parenthesis in~\eqref{eq:marginal} is bounded above by $1$ and maximized when the policy does not experience failure (in which case the conditional and unconditional expectations will be equal). The numerator of the second term bounded $[0, 1]$, while the denominator grows exponentially large when a failure occurs. As a result, the marginal distribution of $\theta^*$ assigns higher probability (relative to the prior) for policies that avoid failure.

\section{Simulation Experiments}\label{experiments}

In this section, we provide empirical comparisons of RADIUM with existing methods for adversarial optimization and policy repair in a range of simulated environments. We have two main goals in this section. First, we seek to understand whether re-framing the failure repair problem from optimization to inference leads to better solutions (i.e. more robust designs and better coverage by the predicted failures). Second, we study whether the gradient-based version of our method yields any benefits over the gradient-free version. After the empirical comparisons in this section, Section~\ref{hw} then demonstrates how well the failure modes and repaired policies generated using our method transfer from simulation to hardware.

We conduct simulation studies on a range of problems from the robotics and cyberphysical systems literature, comparing against previously-published adversarial optimization methods. The code used for our experiments can be found at \projectwebsite{}, and more detail on each benchmark, baseline, and implementation is provided in the supplementary material.

\subsection{Baselines}

We compare with three baselines taken from the adversarial optimization and testing literature.
\textit{Gradient descent with randomized counterexamples (\gdr)} optimizes the design using a fixed set of random counterexamples, representing a generic policy optimization with domain randomization approach.
\textit{Gradient descent with adversarial counterexamples (\gda)} alternates between optimizing the design and optimizing for maximally adversarial failure modes, as in~\cite{hanselmannKINGGeneratingSafetyCritical2022a,dawsonRobustCounterexampleguidedOptimization2022b}.
\textit{Learning to collide (\ltc)} uses black-box optimization (REINFORCE) to search for failure cases~\cite{dingLearningCollideAdaptive2020a}.
We denote the gradient-free and gradient-based variants of RADIUM as $R_0$ and $R_1$, respectively.
All methods were run on the same GPU model with the same sample budget for each task. Hyperparameters for all experiments are given in the appendix.

\subsection{Benchmark problems}

\begin{figure*}[t]
  \centering
  \includegraphics[width=0.8\linewidth]{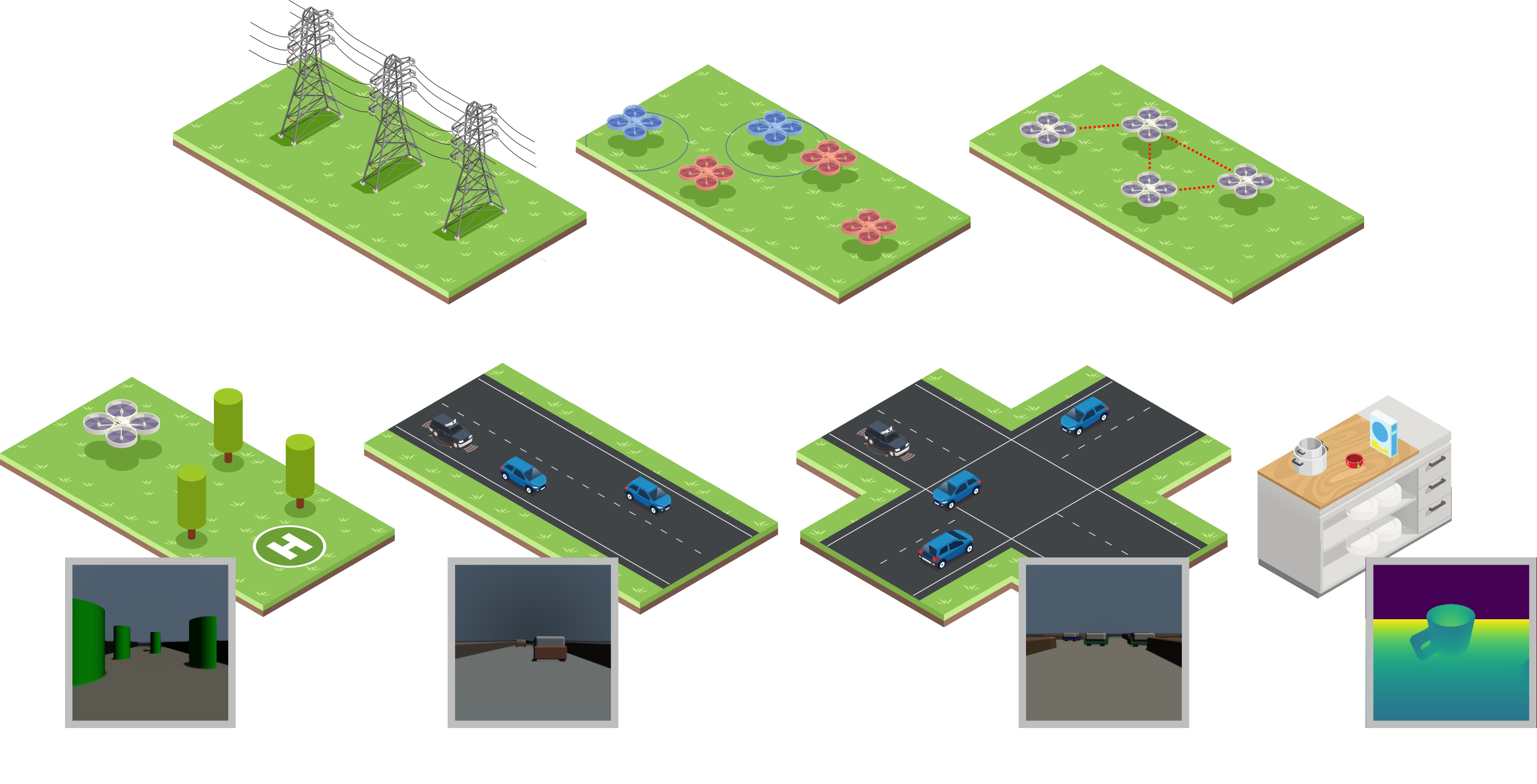}
  \caption{The different environments used in our simulation studies, including 3 domains without visual feedback and 4 domains with vision in the loop.}\label{fig:environments}
\end{figure*}

We rely on two classes of benchmark problem in this work: 3 problems without vision in the loop, and 4 problems with vision in the loop. Within each problem domain, we include multiple environments of varying complexity, for a total of 12 distinct environments. A summary of these environments is given in Fig~\ref{fig:environments}. More details on the parameters and cost functions used for each environment are given in the supplementary material.

\subsubsection{Non-vision benchmarks}
\textbf{Search:} a set of seeker robots must cover a region to detect a set of hiders. $\theta$ and $\phi$ define trajectories for the seekers and hiders, respectively. Failure occurs if any hider escapes detection by the seekers (which have fixed sensing radius). This environment has two variants: small (6 seeker vs. 10 hider, $\dim{\theta} = 60$, $\dim{\phi} = 100$) and large (12 seeker vs. 20 hider, $\dim{\theta} = 120$, $\dim{\phi} = 200$).
\textbf{Formation control:} a swarm of drones fly to a goal while maintaining full connectivity with a limited communication radius. $\theta$ defines trajectories for each robot in the swarm, while $\phi$ defines an uncertain wind velocity field. Failure occurs when the second eigenvalue of the graph Laplacian is close to zero. This environment has small (5 agent, $\dim{\theta} = 30$, $\dim{\phi} = 1280$) and large (10 agent, $\dim{\theta} = 100$, $\dim{\phi} = 1280$) variants.
\textbf{Power grid dispatch:} electric generators must be scheduled to ensure that the network satisfies voltage and maximum power constraints in the event of transmission line outages. $\theta$ specifies generator setpoints and $\phi$ specifies line admittances; failures occur when any of the voltage or power constraints are violated. This environment has small (14-bus, $\dim{\theta} = 32$, $\dim{\phi} = 20$) and large (57-bus, $\dim{\theta} = 98$, $\dim{\phi} = 80$) versions.

\subsubsection{Vision-in-the-loop benchmarks}

\textbf{AV (highway):} An autonomous vehicle must overtake two other vehicles. 
\textbf{AV (intersection):} the autonomous vehicle must navigate an uncontrolled intersection with crossing traffic.
In both AV tasks, the actions of the non-ego vehicles are uncertain, and the AV observes RGBd images from a front-facing camera as well as its own speed.
\textbf{Drone:} A drone must safely navigate through a cluttered environment in windy conditions. There is uncertainty in the wind speed and location of all obstacles.
Initial convolutional neural network (CNN) policies $\theta_0$ for drone and intersection environments were pretrained using behavior cloning, and CNN-based policies for the highway environment were pretrained using PPO~\cite{schulmanProximalPolicyOptimization2017}.
\textbf{Grasp (box/mug): } a robot must locate and grasp an object using a depth image of the scene. There is uncertainty in the location of the objects and in the location of a nearby distractor object. The grasp detector is trained with labels of ground-truth grasp affordances. Fig.~\ref{fig:example_rendering} shows the rendered images for the drone, highway, mug grasping, and intersection tasks (clockwise from top left).

The dimension of the failure space is 20 for the highway task, 30 for the intersection task, 12 for the drone task, and 4 for the grasping tasks. The dimension of the policy space is 64k for the highway and intersection tasks, 84k for the drone task, and 266k for the grasping tasks.

\subsection{Implementation}

Since we require a differentiable renderer and simulation engine for our work, we were not able to use an off-the-shelf simulator like CARLA~\cite{Dosovitskiy17}. Instead, we write our own simulator and basic differentiable renderer using JAX, which is available at \projectwebsite{}.
Likewise, our method and all baselines were implemented in JAX and compiled using JAX's just-in-time (JIT) compilation. Each metric reports the mean and standard deviation across four independent random seeds. All methods are given the same total sample budget for both prediction and repair (except for \gdr, which does not update the predicted failure modes).

The non-vision benchmarks were all initialized with random $\theta_0$, and the vision benchmarks were initialized using $\theta_0$ trained using reinforcement learning or behavior cloning with domain randomization. We include comparisons with \gdr, \gda, and \ltc{} on all non-vision benchmarks. Since $\theta_0$ on the vision benchmarks was trained using domain randomization, \gdr{} is not able to improve the initial parameters, and so we include comparisons with $\theta_0$, \gda, and \ltc{} for the vision benchmarks.

\subsection{Metrics}

To measure the robustness of the optimized policies, we report the failure rate (FR) on a test set of 1,000 i.i.d. samples of $\phi$ from the prior $p_{\phi, 0}$. We also report the mean cost on this test set as well as the maximum cost on the vision-in-the-loop benchmarks (where cost is bounded by construction) and the 99\textsuperscript{th}-percentile cost for non-vision benchmarks (some of which have unbounded cost, making the 99\textsuperscript{th} percentile more representative). Costs are normalized by the maximum cost observed across any method. Finally, for each task, we report the time required for simulating a rollout both with and without reverse-mode differentiation.

\subsection{Results}

\begin{figure*}[t]
  \centering
  \includegraphics[width=\linewidth]{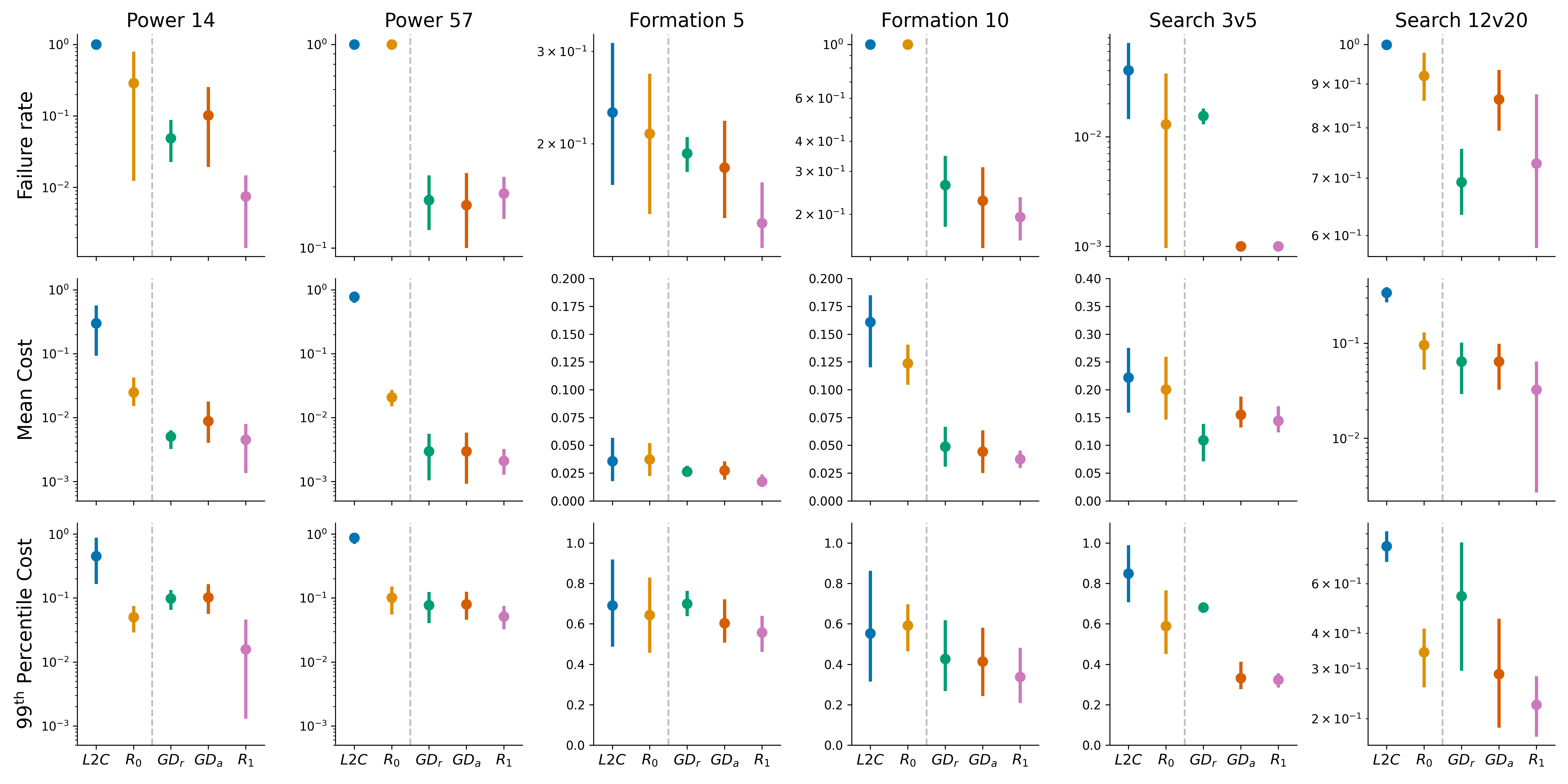}
  \caption{Comparison of our method (gradient-free and gradient-based variants $R_0$ and $R_1$, respectively) and baseline methods on benchmark problems without vision in the loop, showing failure rate, mean cost, and 99\textsuperscript{th} percentile cost on a test set of 1,000 randomly sampled $\phi$. The dashed gray lines separate gradient-free and gradient-based methods.}\label{fig:nonvision_results}
\end{figure*}

\begin{figure*}[t]
  \centering
  \includegraphics[width=\linewidth]{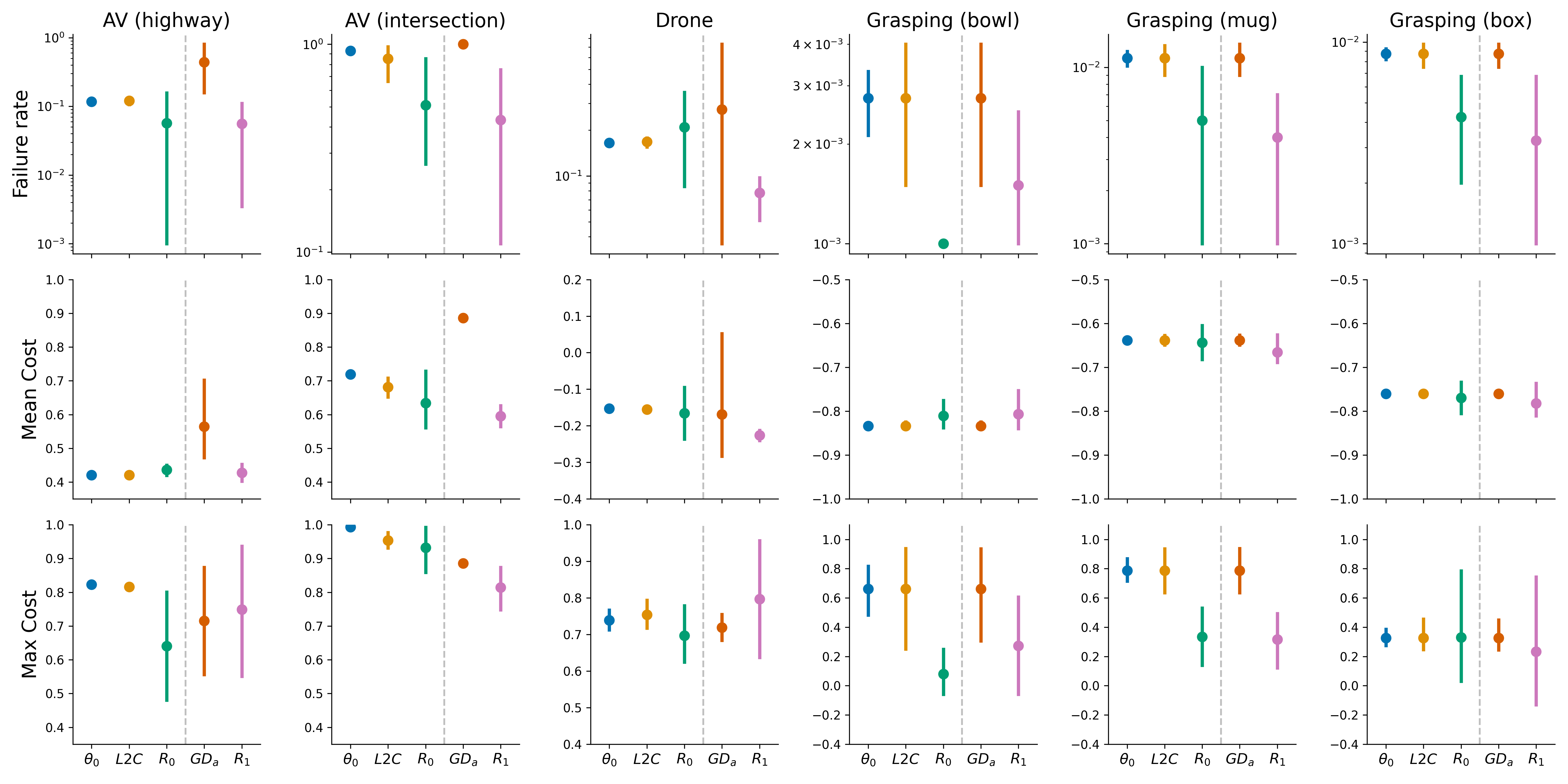}
  \caption{Comparison of our method (gradient-free and gradient-based variants $R_0$ and $R_1$, respectively) and baseline methods on benchmark problems with vision in the loop, showing failure rate, mean cost, and max cost on a test set of 1,000 randomly sampled $\phi$. The dashed gray lines separate gradient-free and gradient-based methods.}\label{fig:vision_results}
\end{figure*}

\begin{figure*}[t]
  \centering
  \includegraphics[width=\linewidth]{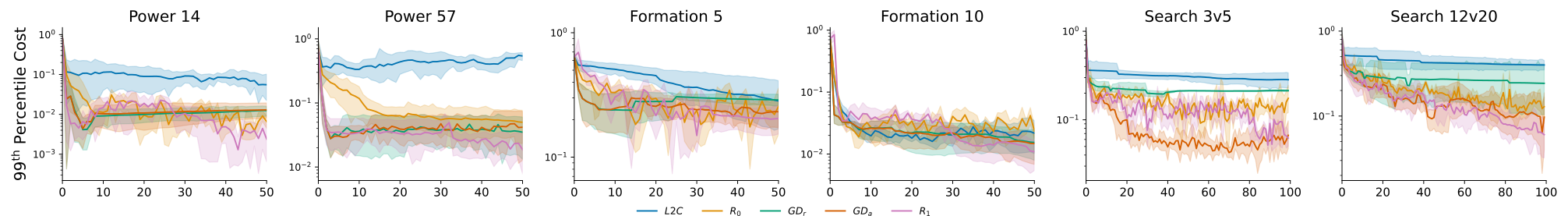}
  \includegraphics[width=\linewidth]{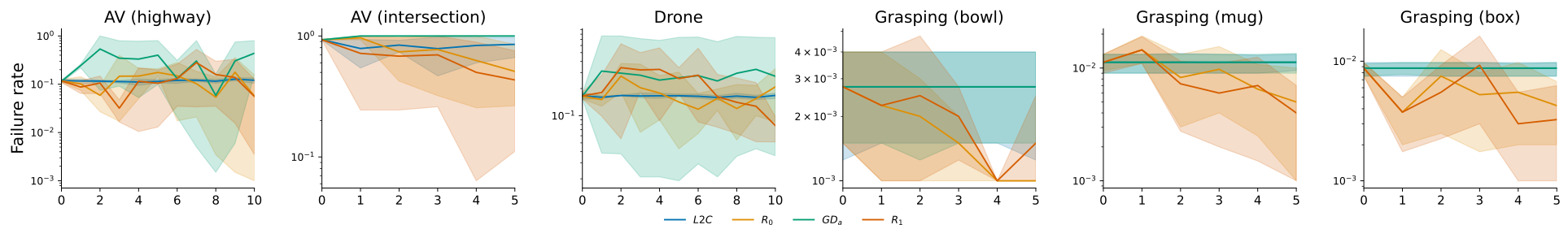}
  \caption{Convergence rates of our method and baselines on tasks with (top) and without (bottom) vision in the loop.}\label{fig:convergence}
\end{figure*}

Fig.~\ref{fig:nonvision_results} shows the results from benchmark problems without vision in the loop, while Fig.~\ref{fig:vision_results} shows the results from problems with vision in the loop. For ease of comparison, each plot groups the gradient-free methods (\ltc{} and $R_0$) and the gradient-based methods (\gdr, \gda, and $R_1$). Since the initial parameters for the vision-in-the-loop benchmarks in Fig.~\ref{fig:vision_results} were trained using RL or behavior cloning (which do not require differentiable simulation), we group $\theta_0$ with the gradient-free results in Fig.~\ref{fig:vision_results}. We also compare the convergence rates of each method in Fig.~\ref{fig:convergence}. Table~\ref{tab:runtime} shows the time required for simulating with and without automatic differentiation.

\begin{figure*}[t]
  \centering
  \includegraphics[width=\linewidth]{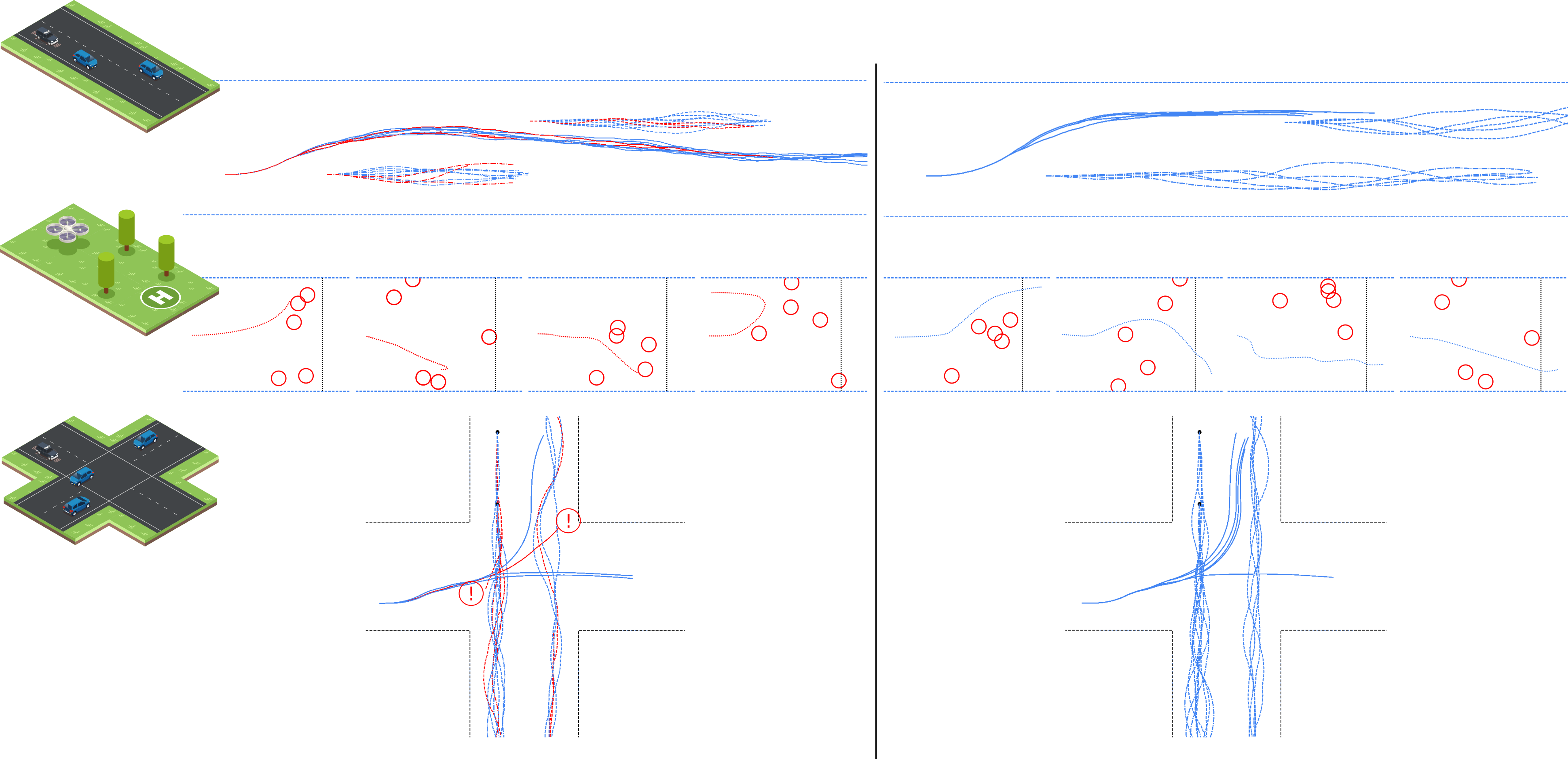}
  \caption{Examples of failure cases (left) and repaired policies (right) generated using our method. Failed trajectories are shown in red.}\label{fig:repairs}
\end{figure*}

Fig.~\ref{fig:repairs} shows examples of failure cases and repaired policies generated using $R_1$ on three vision-in-the-loop tasks: AV (highway), AV (intersection), and drone (static). The left of Fig.~\ref{fig:repairs} shows the initial policy $\theta_0$ and failure modes discovered using our method (sampling $\phi$ while holding $\theta_0$ fixed), while the right shows the repaired policy and updated challenging counterexamples. Since the distribution of failure modes shifts as we repair the policy, we continuously re-sample the failure modes to be relevant to the updated policy. In all cases, we see that the repaired policy found using our method experiences fewer failures, despite the updated adversarial failure modes. In certain cases, the repaired policy exhibits a qualitatively different behavior; for example, in the vision-in-the-loop highway control task, the repaired policy is less aggressive than the original policy, avoiding the risky overtake maneuver (top row of Fig.~\ref{fig:repairs}).

\begin{table*}
\caption{Time required for simulating a rollout with and without autodiff (AD) for each task (in seconds). Average and standard deviation (subscript) reported across 100 trials on AMD Ryzen Threadripper 3990X 64-Core Processor (non-vision tasks) and an NVIDIA RTX A4000 (vision-in-the-loop tasks).}
\label{tab:runtime}
\centering
\begin{tabular}{rcccccc}
\toprule
\multicolumn{1}{l}{} & \multicolumn{6}{c}{Non-vision tasks}                                        \\
\cmidrule(r){2-7}
                     & Power (14)         & Power (57)             & Formation (5)    & Formation (10) & Search (3v5)                  & Search (12v20) \\
\midrule
Without AD           & $0.00122_{0.00413}$   & $0.0107_{0.00893}$ & $0.0326_{0.0173}$ & $0.628_{0.296}$ & $0.00147_{0.00461}$ & $0.00461_{0.00747}$ \\
With AD              & $0.00165_{0.00488}$   & $0.0136_{0.0111}$  & $0.0543_{0.0212}$ & $0.714_{0.306}$ & $0.00358_{0.00704}$ & $0.0107_{0.00851}$ \\
\midrule
\multicolumn{1}{l}{} & \multicolumn{6}{c}{Vision-in-the-loop}                                        \\
\cmidrule(r){2-7}
                     & AV (hw.)           & AV (int.)              & Drone (st.)      & Drone (dyn.)    & Grasp (all) & \\
\midrule
Without AD           & $0.70_{0.003}$     & $2.22_{0.01}$          & $0.39_{0.002}$   & $0.39_{0.001}$  & $0.0045_{5.1\times 10^{-5}}$ &  \\
With AD              & $1.72_{0.003}$     & $6.65_{0.14}$          & $1.77_{0.06}$    & $1.83_{0.04}$   & $0.0049_{3.8\times 10^{-5}}$ & \\
\bottomrule
\end{tabular}
\end{table*}

\subsection{Discussion}

In our results on problems without vision in the loop (Fig.~\ref{fig:nonvision_results}), we see several high-level trends. First, we see that gradient-based techniques (\gdr, \gda, and $R_1$) achieve lower failure rates, mean cost, and 99\textsuperscript{th} percentile costs relative to gradient-free methods (\ltc{} and $R_0$) on the test set, likely because gradient information helps the former methods explore the high-dimensional search space (as seen in the faster convergence of gradient-based methods on problems without vision in Fig.~\ref{fig:convergence}). Moreover, we find that our methods ($R_0$ and $R_1$) outperform other methods within each of their respective categories; i.e. $R_0$ yields repaired solutions with lower costs and failure rates than \ltc, and $R_1$ likewise outperforms \gdr{} and \gda{}.

We see a slightly different pattern in our results for problems with vision in the loop. On these problems, we find that existing gradient-based methods like \gda{} do not achieve lower failure rates than gradient-free methods like \ltc{}, possibly due to poor gradient quality from the differentiable renderer (where occlusions can lead to large variance in the automatically-derived gradients). In contrast, both variants of our method achieve low failure rates for repaired policies in the vision-in-the-loop tasks, and $R_1$ in particular is able to achieve better performance on some tasks because the Metropolis-Hastings adjustment on line~\ref{mala:mh} of Algorithm~\ref{alg:mala} allows it to reject large steps caused by poorly conditioned gradients.

\section{Hardware demonstration} \label{hw}

To demonstrate that the repaired policies found using our method can be successfully transferred to hardware, we present results from three sets of hardware experiments. Our primary hardware demonstration in Section~\ref{hw:car} involves predicting and repairing failure modes for a vision-in-the-loop driving policy on a 1/10th-scale race car, where we show not only that failures predicted in simulation transfer to reality, but also that the repaired policy is also able to avoid these failures in reality, despite being trained only on simulated failures. We also include results in Section~\ref{hw:other} demonstrating sim2real transfer of non-vision based policies for the multi-agent search problem.

\subsection{Sim2real transfer of repaired vision-in-the-loop policies}\label{hw:car}

We pre-train a policy for the highway task that has three components: a tracking controller that follows a pre-planned trajectory, a model-based collision avoidance controller that attempts to avoid rear-ending cars in front of the ego vehicle (using the depth camera to measure the distance to the next car), and a neural network controller that accelerates and steers based on the depth image received from a forward facing camera. The parameters of the neural network and the pre-planned trajectory are optimized via vanilla gradient descent during pre-training.

\begin{figure*}[tb]
\centering
\subfloat[Nominal policy]{
    \includegraphics[height=6cm]{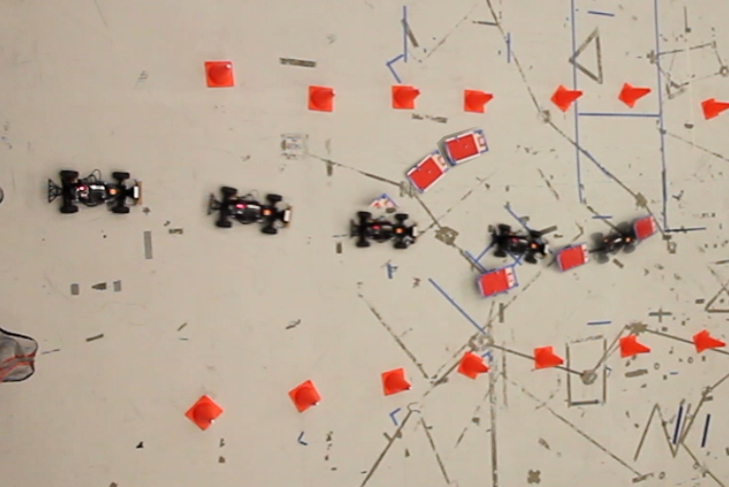}
    \label{fig:hw:car:fail}
}
\subfloat[Repaired policy]{
    \includegraphics[height=6cm]{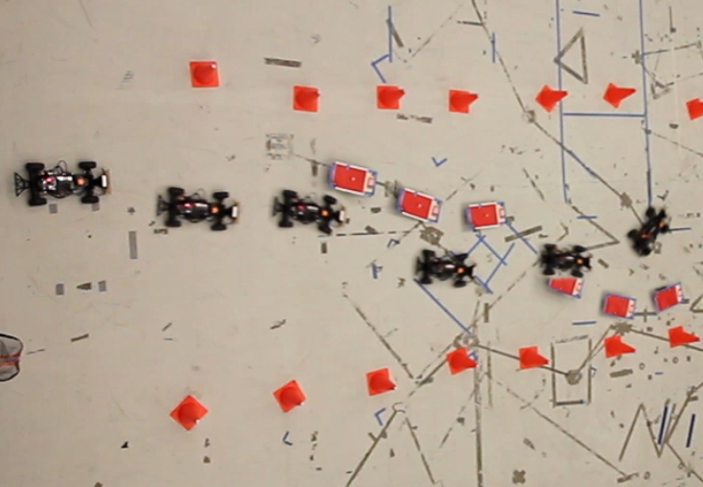}
    \label{fig:hw:car:repair}
}
\caption{Composite images from hardware experiments with vision-in-the-loop controllers. (Left) the nominal policy and the worst adversarial example found using our method, where a crash occurs. (Right) the repaired policy and new worst adversarial example (both found using our method); the repaired policy avoids crashing in this case. Videos are included in the supplementary materials.}
\label{fig:hw:car}
\end{figure*}

We then use $R_1$ to predict failure modes for this pre-trained, vision-in-the-loop policy in simulation, then transfer both the pre-trained policy and predicted failure modes to hardware. As shown in Fig.~\ref{fig:hw:car:fail}, the failure modes predicted in simulation correspond to real failures on hardware. We then repair the policy using $R_1$, which results in an updated policy and new predicted worst-case failure mode, shown in Fig.~\ref{fig:hw:car:repair}. Since $R_1$ predicts different failure modes for the nominal and repaired policies, Table~\ref{tab:hw:metrics} compares the failure rates of both policies on 20 independent samples of $\phi$ from the prior distribution, showing that the repaired policy is 5x safer than the original policy.

These results demonstrate that both the predicted failure modes and the repaired policy can successfully transfer from simulation to hardware, despite the gap between the simulated dynamics and rendering system and reality. A benefit of the sampling-based approach proposed in $R_0$ and $R_1$ is that the noise added during the sampling step helps us avoid converging to narrow local minima, avoiding failures that occur only due to quirks in the simulation environment.

\input{tables/hw_metrics}

\subsection{Sim2real transfer of non-vision-based policies}\label{hw:other}

\begin{figure*}[tb]
    \centering
    \includegraphics[width=0.7\linewidth]{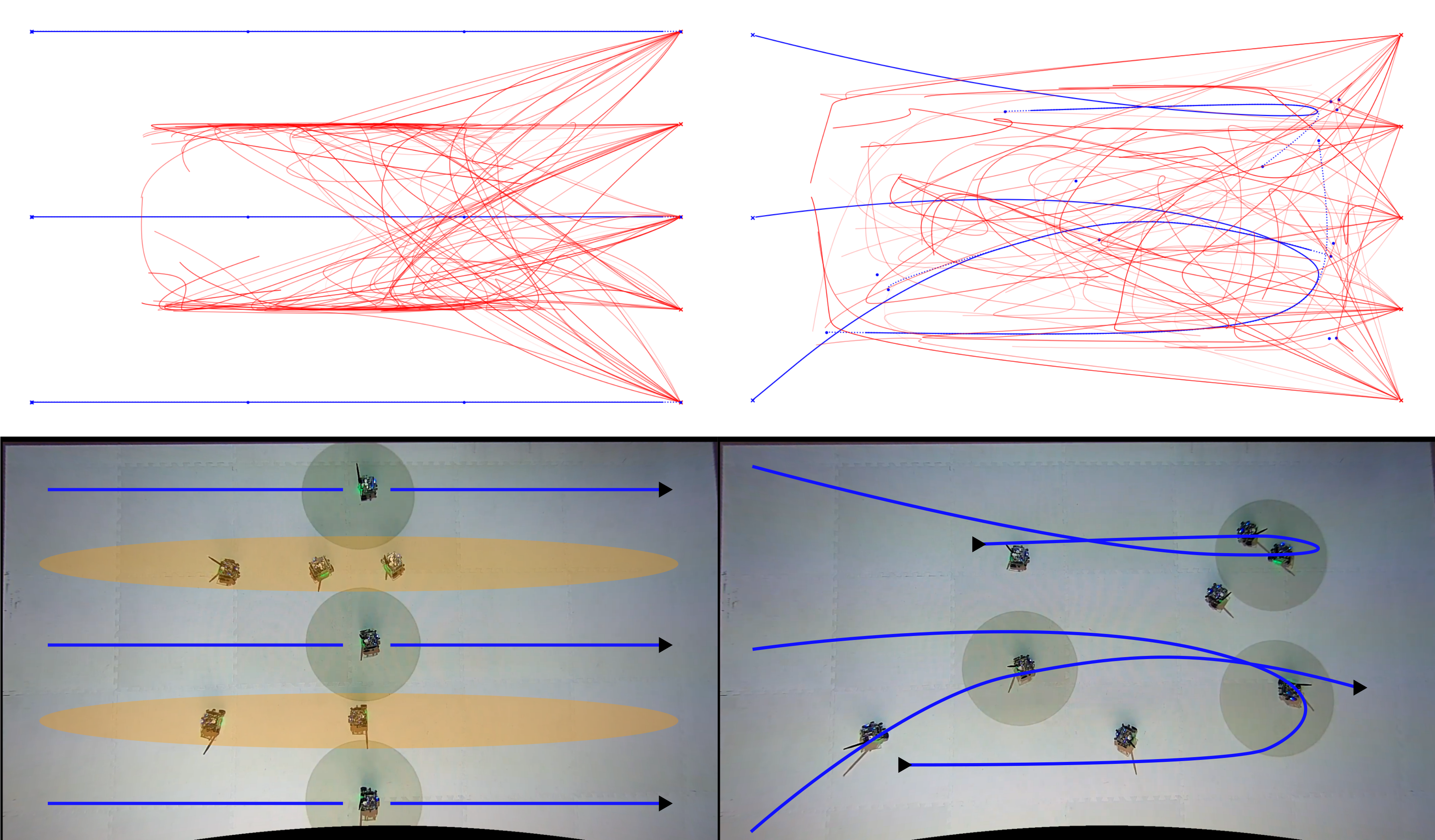}
    \caption{(Left) HW results for search-evasion with 5 hiders and 3 seekers, showing an initial search pattern (blue) and predicted failure modes (red). (Right) HW results for an optimized search pattern leaves fewer hiding places.}
    \label{fig:hw_experimental_results}
\end{figure*}

We deploy the optimized hider and seeker trajectories in hardware using the Robotarium multi-robot platform~\cite{wilsonRobotariumGloballyImpactful2020a} (we use 3 seekers and 5 hiders, since we had difficulty testing with more agents in the limited space). We first hold the search pattern (design parameters) constant and optimize evasion patterns against this fixed search pattern, yielding the results shown on the left in Fig.~\ref{fig:hw_experimental_results} where the hiders easily evade the seekers. We then optimize the search patterns using our approach, yielding the results on the left where the hiders are not able to evade the seekers.

\section{Conclusion}

In this paper, we have proposed a novel framework for predicting the ways in which a learning-based system might fail and repairing the learned policy to preemptively mitigate those failures. Our framework reframes traditional adversarial optimization as an iterative sampling process to prioritize diversity in the predicted failures, yielding more robust repaired policies and avoiding overfitting to a narrow set of predicted failures. We present both gradient-free and gradient-based variants of our framework, and we examine the tradeoff in computation time and solution quality between these methods, using end-to-end gradients through differentiable simulation and rendering for the gradient-based variant. We find that our method yields more robust repaired policies than prior methods on a range of problems, and that both the predicted failure modes and repaired policies transfer from simulation to reality.

\subsection{Limitations \& future work}

The main limitation of our approach is that it requires a simulator with enough fidelity to predict the failures of interest; moreover, the gradient-based version of our framework requires a differentiable simulator, which can be even more difficult to come by. This means that our method cannot predict any failures not modeled by the simulator (e.g. hardware issues like loose cables or software issues like dropped packets). In future, we hope to explore algorithms for combining accelerated testing in simulation with limited higher fidelity testing in hardware.

In future work, we also hope to integrate with emerging photorealistic differentiable renderers like Mitsuba~\cite{jakobDRJITJustintime2022,Mitsuba3}. As of this writing, ease of use and interoperability with machine learning libraries is still a challenge for these renderers, but we hope that additional engineering effort in this area will result in an easy-to-use, high-quality differentiable renderer that can be integrated with existing robotics simulation frameworks.

\appendices

\section{Proof of Theorem~\ref{thm:convergence}}

We will show the proof for sampling from the failure generating process with likelihood given by Eq.~\eqref{eq:failure_logprob_tempered}; the proof for the repair generating process follows similarly. The log-likelihood for the failure generating process is
\begin{equation}
    \log p_{\phi, 0}(\phi) - \tau [J^* - J(\theta, \phi)]_+ \label{eq:failure_logprob}
\end{equation}

\cite{maSamplingCanBe2019} show that MALA sampling enjoys the convergence guarantees of Theorem~\ref{thm:convergence} so long as the target log likelihood is strongly convex outside of a ball of finite radius $R$ (see Theorem 1 in \cite{maSamplingCanBe2019}). Since $\log p_{\phi, 0}(\phi)$ is assumed to be strongly $m$-convex, it is sufficient to show that as $||\phi|| \to \infty$, the strong convexity of the log-prior dominates the non-convexity in $\tau [J^* - J(\theta, \phi)]_+$.

For convenience, denote $f(\phi) = -\tau [J^* - J(\theta, \phi)]_+$ and $g(\phi) = \log p_{\phi, 0}(\phi)$. We must first show that $f(\phi) + g(\phi)$ is $(m-L)$-strongly convex, for which it suffices to show that $f(\phi) + g(\phi) - (m-L)/2 ||\phi||^2$ is convex. Note that
\begin{align}
    f(\phi) + g(\phi) - \frac{m-L}{2} ||\phi||^2 &= f(\phi) + \frac{L}{2} ||\phi||^2 + g(\phi) - \frac{m}{2} ||\phi||^2
\end{align}
$g(\phi) - \frac{m}{2} ||\phi||^2$ is convex by $m$-stong convexity of $g$, so we must show that the remaining term, $f(\phi) + L/2 ||\phi||^2$, is convex. Note that the Hessian of this term is $\nabla^2 f(\phi) + LI$. Since we have assumed that $J$ is $L$-Lipschitz smooth (i.e. its gradients are $L$-Lipschitz continuous), it follows that the magnitudes of the eigenvalues of $\nabla^2 f$ are bounded by $L$, which is sufficient for $\nabla^2 f(\phi) + LI$ to be positive semi-definite, completing the proof.

\section{Proof of Theorem~\ref{thm:joint}}

We can treat Alg.~\ref{alg:adv_diffusion} as a two-stage Gibbs sampling procedure and apply the Hammersley-Clifford Theorem~\cite{robertMonteCarloStatistical2004} to get the joint distribution
\begin{equation*}
    f_{\theta, \phi}(\theta^*, \phi^*) = p_{\theta, 0}(\theta^*) p_{\phi, 0}(\phi^*) \frac{e^{-[J^* - J(\theta^*, \phi^*)]_+}}{\expectation_{\phi \sim p_{\phi, 0}}\left[ e^{J(\theta^*, \phi) - J^*} \right]} \label{eq:joint}
\end{equation*}

Integrating over $\phi$ yields the marginal distribution of $\theta$ in Equations~\eqref{eq:integration_start}--\eqref{eq:integration_end}, completing the proof.
\begin{figure*}[!b]
\normalsize

\vspace*{4pt}
\hrulefill

\begin{align}
    f_{\theta^*} &= \int_\phi f_{\theta, \phi}(\theta^*, \phi) d\phi = \frac{p_{\theta, 0}(\theta^*)}{\expectation_{\phi \sim p_{\phi, 0}}\left[ e^{J(\theta^*, \phi) - J^*} \right]}  \int_\phi p_{\phi, 0}(\phi) e^{-[J^* - J(\theta^*, \phi^*)]_+} d\phi \label{eq:integration_start}\\
    &= p_{\theta, 0}(\theta^*)\frac{\expectation_{\phi \sim p_{\phi, 0}}\left[ e^{-[J^* - J(\theta^*, \phi)]_+} \right]}{\expectation_{\phi \sim p_{\phi, 0}}\left[ e^{J(\theta^*, \phi) - J^*} \right]} \\
    &= p_{\theta, 0}(\theta^*)\frac{\expectation_{\phi \sim p_{\phi, 0}}\left[ e^{-(J^* - J(\theta^*, \phi))} | J^* - J(\theta^*, \phi) \geq 0 \right] + \expectation_{\phi \sim p_{\phi, 0}}\left[ 1 | J^* - J(\theta^*, \phi) < 0 \right]}{\expectation_{\phi \sim p_{\phi, 0}}\left[ e^{J(\theta^*, \phi) - J^*} \right]} \\
    &= p_{\theta, 0}(\theta^*)\frac{\expectation_{\phi \sim p_{\phi, 0}}\left[ e^{-(J^* - J(\theta^*, \phi))} | J^* \geq J(\theta^*, \phi) \right] + \expectation_{\phi \sim p_{\phi, 0}}\left[ 1 | J^* < J(\theta^*, \phi) \right]}{\expectation_{\phi \sim p_{\phi, 0}}\left[ e^{J(\theta^*, \phi) - J^*} \right]} \\
    &= p_{\theta, 0}(\theta^*)\frac{\expectation_{\phi \sim p_{\phi, 0}}\left[ e^{-(J^* - J(\theta^*, \phi))} | J^* \geq J(\theta^*, \phi) \right] + \mathbb{P}[J(\theta^*, \phi) > J^*]}{\expectation_{\phi \sim p_{\phi, 0}}\left[ e^{J(\theta^*, \phi) - J^*} \right]} \label{eq:integration_end}
\end{align}
\end{figure*}

\section{Details on experiments}

Tables~\ref{tab:hyperparams_nonvision} and~\ref{tab:hyperparams_vision} include the hyperparameters used for each task. All methods use the same hyperparameters (except that only $R_1$ includes quenching).

\begin{table}[tb]
    \caption{Hyperparameters used for each non-vision environment. $n_{q}$ is the number of quenching rounds. \label{tab:hyperparams_nonvision}}
    \begin{tabular}{llllllll}
        \hline
        Environment    & $J^*$ & $n$ & $\epsilon$             & $K$ & $M$ & $n_q$ & $\alpha$ \\ \hline
        Formation (5)  & 10.0  & 5   & $10^{-5}$              & 50  & 50  & 20    & $5$      \\
        Formation (10) & 10.0  & 5   & $10^{-4}$              & 50  & 50  & 20    & $5$      \\
        Search (3v5)   & -0.1  & 10  & $10^{-2}$              & 50  & 50  & 20    & $5$      \\
        Search (12v20) & -0.1  & 10  & $10^{-2}$              & 50  & 50  & 20    & $5$      \\
        Power (14)     & 4.0   & 10  & $10^{-6}$ for $\theta$ & 10  & 50  & 25    & $5$      \\
                       &       &     & $10^{-2}$ for $\phi$   &     &     &       &          \\
        Power (57)     & 6.0   & 10  & $10^{-6}$ for $\theta$ & 10  & 50  & 20    & $5$      \\
                       &       &     & $10^{-4}$ for $\phi$   &     &     &       &          \\
        \hline
    \end{tabular}
\end{table}

\begin{table}[b]
\caption{Hyperparameters used for each vision environment. $n_{q}$ is the number of quenching rounds. \label{tab:hyperparams_vision}}
\begin{tabular}{lllllll}
\hline
Environment                             & $n$ & $\epsilon$            & $K$ & $N$ & $n_q$ & $\alpha$ \\ \hline
Formation (5)                    & 5       & $10^{-3}$         & 50  & 5   & 5             & $5$ \\
Formation (10)                   & 5       & $10^{-3}$         & 50  & 5   & 5             & $5$ \\
Search (3v5)  & 10     & $10^{-3}$         & 100 & 10  & 25            & $5$ \\
Search (12v20) & 10    & $10^{-3}$         & 100 & 10  & 25            & $5$ \\
Power (14)                     & 10     & $10^{-6}$ for $\theta$ & 100 & 10  & 10            & $5$ \\
                                        &       & $10^{-2}$ for $\phi$ &     &     &    &           \\
Power (57)                     & 10     & $10^{-6}$ for $\theta$ & 100 & 10  & 10            & $5$ \\
                                        &       & $10^{-2}$ for $\phi$ &     &     &      &        \\
                                        \hline
\end{tabular}
\end{table}

\section*{Acknowledgment}

C. Dawson is supported by the NSF GRFP under Grant No. 1745302. This work was partly supported by the National Aeronautics and Space Administration (NASA) ULI grant 80NSSC22M0070, Air Force Office of Scientific Research (AFOSR) grant FA9550-23-1-0099, and the Defense Science and Technology Agency in Singapore. Any opinions, findings, and conclusions or recommendations expressed in this publication are those of the authors and do not necessarily reflect the views of the sponsors.

\ifCLASSOPTIONcaptionsoff
  \newpage
\fi


\bibliographystyle{IEEEtran}
\bibliography{IEEEabrv,main}

%

\begin{IEEEbiography}[{\includegraphics[width=1in,height=1.25in,clip,keepaspectratio]{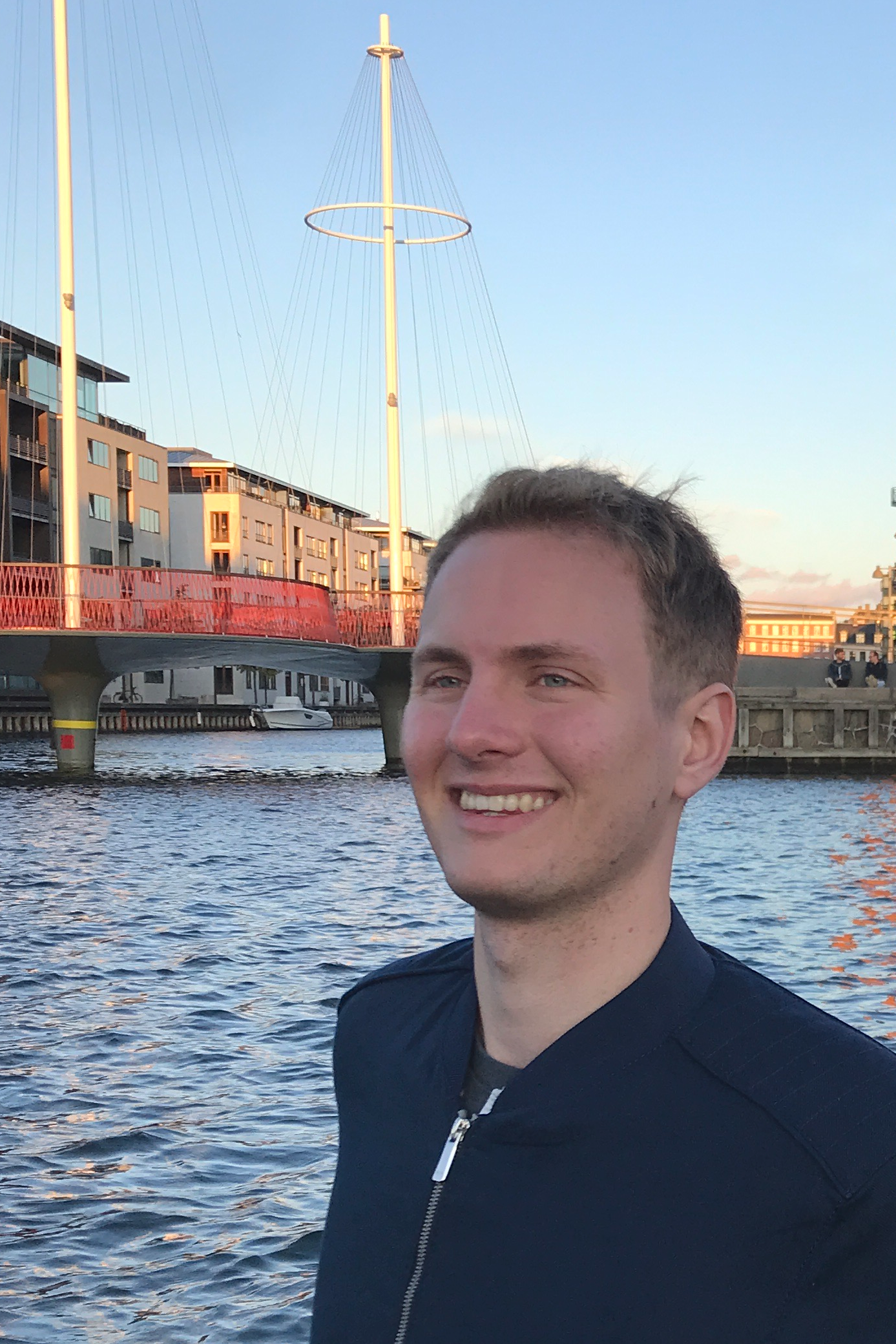}}]{Charles Dawson}
is a graduate student in the Department of Aeronautics and Astronautics at MIT, supported by the NSF Graduate Research Fellowship. He works on using tools from controls, learning, and optimization to understand safety in autonomous and cyberphysical systems. Prior to MIT, he received a B.S. in Engineering from Harvey Mudd College.
\end{IEEEbiography}

\begin{IEEEbiography}[{\includegraphics[width=1in,height=1.25in,clip,keepaspectratio]{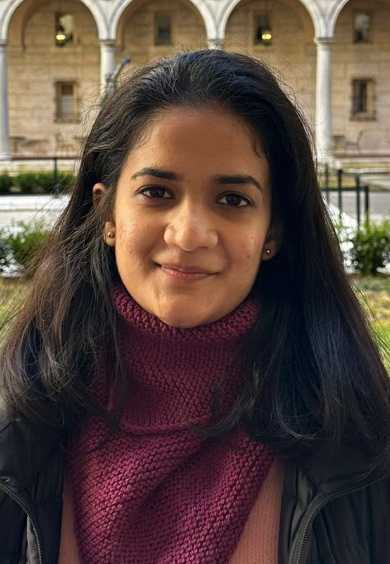}}]{Anjali Parashar}
Anjali Parashar received the B.Tech in Mechanical Engineering degree from Indian Institute of Technology, Indore (IITI), India in 2020 and the S.M. in Mechanical Engineering from Massachusetts Institute of Technology (MIT), Cambridge, MA, USA in 2023 where she is currently working towards the Ph.D. degree with the Department of Mechanical Engineering. She is a member of the Reliable Autonomous Systems Lab, led by Prof. Chuchu Fan. Her research interests include robotics and optimization with applications to verification of autonomous systems
\end{IEEEbiography}

\begin{IEEEbiography}[{\includegraphics[width=1in,height=1.25in,clip,keepaspectratio]{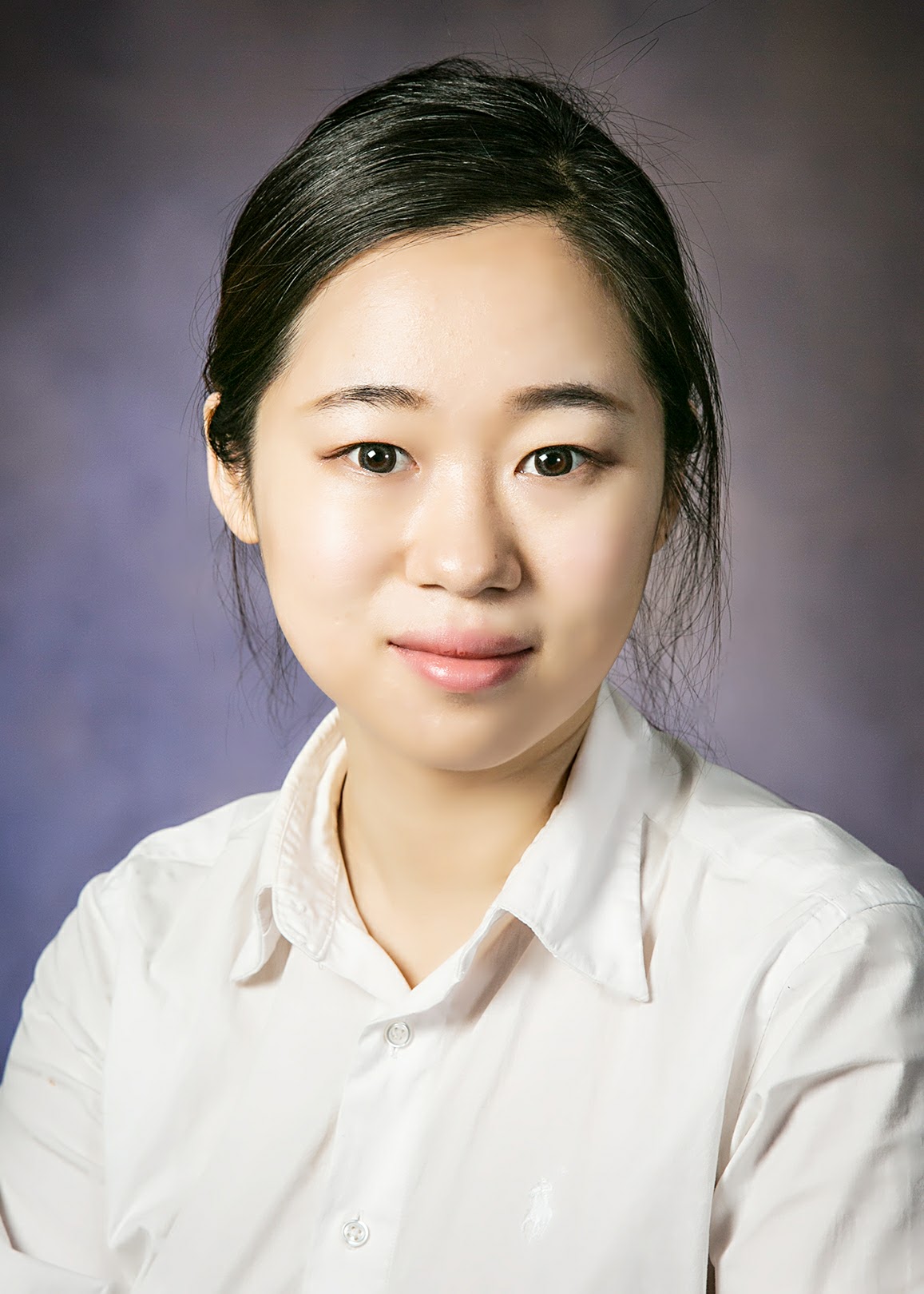}}]{Chuchu Fan}
is an Assistant Professor in AeroAstro and LIDS at MIT. Before that, she was a postdoctoral researcher at Caltech and got her Ph.D. from the Electrical and Computer Engineering Department at the University of Illinois at Urbana-Champaign. Her group at MIT, REALM, works on using rigorous mathematics, including formal methods, machine learning, and control theory, to design, analyze, and verify safe autonomous systems. She is the winner of the NSF CAREER Award, AFSOR YIP Award, Innovator under 35 by MIT Technology Review, and the ACM Doctoral Dissertation Award.
\end{IEEEbiography}





\end{document}

%% file: tables/hw_metrics.tex
\begin{table}[htb]
    \caption{
        Failure rate on 1000 simulated and 20 hardware trials of nominal and repaired policies with exogenous parameters sampled i.i.d. from the prior $p_{\phi, 0}$.
    }
    \label{tab:hw:metrics}
    \begin{center}
        \begin{tabular}{lcc}
            \toprule
            Policy   & Failure rate (simulation) & Failure rate (hardware) \\
            \midrule
            Nominal  & 4.4 \%                    & 25\%                    \\
            Repaired & 0.6 \%                    & 5\%                     \\
            \bottomrule
        \end{tabular}
    \end{center}
\end{table}